\documentclass[11pt]{article}
\usepackage{amsmath,amssymb,url,fullpage}
\usepackage{amsthm}
\usepackage{graphicx}

\def\KL{\mathrm{KL}}
\def\Bhat{\mathrm{Bhat}}
\def\sJ{\mathrm{sJ}}
\def\JB{\mathrm{JB}}
\def\dx{\mathrm{d}x}
\def\dnu{\mathrm{d}\nu}
\def\co{\mathrm{co}}

\def\calC{\mathcal{C}}
\def\calP{\mathcal{P}} 
\def\calN{\mathcal{N}} 
\def\calX{\mathcal{X}}

\def\bbR{\mathbb{R}} 
\def\bbP{\mathbb{P}}

\title{A generalization of the Jensen divergence: The chord gap divergence}
  
\author{Frank Nielsen\footnote{Contact e-mail: {\tt Frank.Nielsen@acm.org} 
}}
 
\date{}

\begin{document}
 
\maketitle

\begin{abstract}
We introduce a novel family of distances, called the chord gap divergences, that generalizes the Jensen divergences (also called the Burbea-Rao distances), and study its properties. It follows a generalization of the celebrated statistical Bhattacharyya distance that is frequently met in applications.
We  report an iterative concave-convex procedure for computing centroids, and analyze the performance of the $k$-means++ clustering with respect to that new dissimilarity measure by introducing the Taylor-Lagrange remainder form of the skew Jensen divergences.
\end{abstract}
\noindent{\bf Key words}
Jensen divergence, Burbea-Rao divergence, Bregman divergence, Jensen-Bregman divergence, Bhattacharrya distance, Kullback-Leibler divergence, centroid, $k$-means++.
 
\section{Introduction}
\label{sec:intro}

In many applications, one faces the crucial dilemma of choosing an appropriate distance $D(\cdot,\cdot)$ between data elements.
In some cases, those distances can be picked up {\em a priori} from well-grounded principles (e.g., Kullback-Leibler distance in statistical estimation~\cite{CT-2012}). 
In other cases, one is rather left at testing several distances~\cite{Bhat-1982}, and choose {\em a posteriori} the  distance 
that yielded the best performance. 
For the latter cases, it is judicious to consider a family of parametric distances $D_\alpha(\cdot,\cdot)$, and learn~\cite{learningalpha-2010} the hyperparameter $\alpha$ according to the application at hand and potentially the dataset (distance selection).
Thus it is interesting to consider parametric generalizations of common distances~\cite{Dico-2009} to improve performance in applications.

Some distances can be designed from inequality gaps~\cite{Jensen-1982,Holder-2017}.
For example, the Jensen divergence $J_F(p,q)$ (also called the Burbea-Rao divergence~\cite{Jensen-1982,Jensen-2011}) is designed from the inequality gap of Jensen inequality
\begin{equation}
F\left(\frac{p+q}{2}\right) \leq \frac{F(p)+F(q)}{2},
\end{equation}
 that holds for a strictly real-valued convex function $F$: 
\begin{equation}
J_F(p,q)=\frac{F(p)+F(q)}{2}-F\left(\frac{p+q}{2}\right).
\end{equation}
We can extend the Jensen divergence to a parametric family of skew Jensen divergences $J_F^\alpha$ (with $\alpha\in (0,1)$) built on the convex inequality gap:
\begin{equation}
F((1-\alpha)p+\alpha q) \leq (1-\alpha)F(p)+\alpha F(q).
\end{equation}
The skew Jensen divergences $J_F^\alpha$ are defined by:
\begin{equation}\label{eq:jf}
J_F^\alpha(p:q)=(1-\alpha)F(p)+\alpha F(q)-F((1-\alpha)p+\alpha q), 
\end{equation}
satisfying $J_F^\alpha(q:p)=J_F^{1-\alpha}(p:q)$ and $J_F^{\frac{1}{2}}(p:q)=J_F(p,q)$.
Here the ':' notation emphasizes the fact that the distance is potentially asymmetric: $J_F^\alpha(p:q)\not =J_F^\alpha(q:p)$.
The term divergence is used in information geometry~\cite{IG-2016} to refer to the smoothness property of the distance function that yields an information-geometric structure of the space induced by the divergence.
Let $[p,q]=\{(pq)_\lambda:=(1-\lambda)p+\lambda q, \lambda\in[0,1]\}$ denote the line segment with endpoints $p$ and $q$.
For scalars $a$ and $b$, $[a,b]$ denotes the interval $[\min(a,b),\max(a,b)]$.
Then we can rewrite Eq.\ref{eq:jf}:
\begin{equation}
J_F^\alpha(p:q)=(F(p)F(q))_\alpha-F((pq)_\alpha).
\end{equation}
In applications, it is rather the relative comparisons of distances rather than their absolute values that is important.
Thus we may multiply a distance by any positive scaling factor and include it in the class of that distance. 
When $F$ is strictly convex and differentiable, the class of Jensen divergences include in the limit cases the Bregman divergences~\cite{Jensen-2011,Zhang-2004,Bregman-2005}:
\begin{eqnarray}
\lim_{\alpha\rightarrow 0^+} \frac{J_F^\alpha(p:q)}{\alpha}&=&B_F(q:p),\\
\lim_{\alpha\rightarrow 1^-} \frac{ J_F^\alpha(p:q)}{1-\alpha}&=&B_F(p:q),
\end{eqnarray}
where 
\begin{equation}
B_F(p:q)=F(p)-F(q)-(p-q)^\top\nabla F(q),
\end{equation}
is the Bregman divergence~\cite{JB-2013}. 
Overall, one may define the smooth parametric family of scaled skew Jensen divergences: 
\begin{equation}
\sJ_F^\alpha(p:q) =\frac{1}{\alpha(1-\alpha)}J_F(p:q),
\end{equation}
that encompasses the Bregman divergence $B_F(p:q)$ and the reverse Bregman divergence $B_F(q:p)$ in limit cases (with $\alpha\in\bbR$).

\begin{figure}
	\centering
	\includegraphics[width=\textwidth]{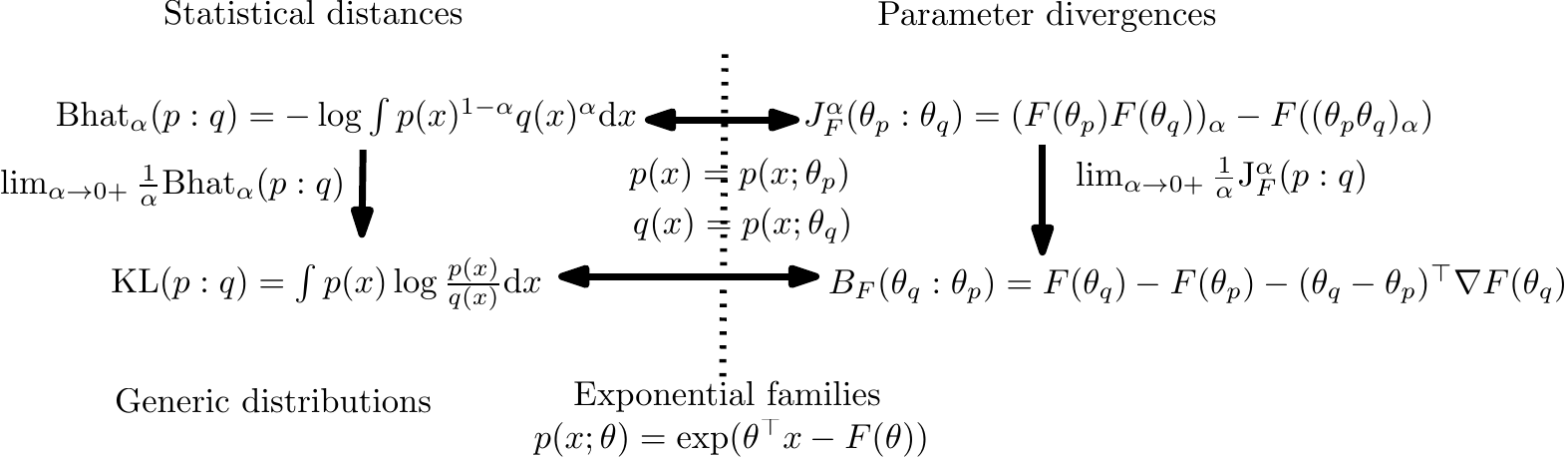}
	\caption{Links between the statistical skew Bhattacharyya distances and parametric skew Jensen divergences when distributions belong to the same exponential family.
	\label{fig:link}}
\end{figure}

There is a nice relationship between the  Jensen divergences operating on parameters (e.g., vectors, matrices) and a class of statistical distances between probability distributions (see Figure~\ref{fig:link}):
Let $\{p(x;\theta)\}_\theta$ be an exponential family~\cite{Bregman-2005} (includes the Gaussian family and the finite discrete ``multinoulli'' family) with convex cumulant function $F(\theta)$.
Then the skew Bhattacharryya distance~\cite{Bhat-1946}:
\begin{eqnarray}
\Bhat_\alpha(p:q)=-\log \int p(x)^{1-\alpha}q(x)^\alpha\dx,\\
\Bhat_\alpha(p:q)=-\log \int p(x)\left(\frac{q(x)}{p(x)}\right)^{\alpha}\dx,
\end{eqnarray} 
between two distributions  belonging to the same exponential family amounts to a skew Jensen divergence~\cite{Jensen-2011}:
\begin{equation}\label{eq:bhat2jensen}
\Bhat(p(x;\theta_1):p(x;\theta_2))= J_F^\alpha(\theta_1:\theta_2).\\
\end{equation}
We further check that:
\begin{eqnarray}
\lim_{\alpha\rightarrow 0^+} \frac{1}{\alpha}\Bhat_\alpha(p:q)&=&\KL(p:q),\\
\lim_{\alpha\rightarrow 1^-} \frac{1}{1-\alpha}\Bhat_\alpha(p:q)&=&\KL(q:p),
\end{eqnarray} 
where
\begin{equation}
\KL(p:q)=\int p(x)\log\frac{p(x)}{q(x)}\dx,
\end{equation} 
is the Kullback-Leibler divergence~\cite{CT-2012}.

The proof of the skewed Bhattacharrya distance converging to the Kullback-Leibler divergence~\cite{CT-2012} proceeds as follows:
We have:
\begin{equation}
\left(\frac{q(x)}{p(x)}\right)^\alpha=\exp\left(\alpha\log \frac{q(x)}{p(x)}\right)\simeq_{\alpha\rightarrow 0} 1+\alpha\log \frac{q(x)}{p(x)}.
\end{equation}
Thus we get:
\begin{equation}
\log p(x)^{1-\alpha}q(x)^{\alpha}  \simeq_{\alpha\rightarrow 0} \log \int \left(p(x)+\alpha\log \frac{q(x)}{p(x)}\right)\dx=\log (1-\alpha\KL(p:q)),
\end{equation}
 and therefore 
\begin{equation}
\lim_{\alpha\rightarrow 0^+} \frac{1}{\alpha}\Bhat_\alpha(p:q)=\KL(p:q),
\end{equation} 
since $\log(1+x)\simeq x$ when $x\rightarrow 0$.

In statistical signal processing, information fusion and machine learning, one often considers the skew  Bhattacharryya distance~\cite{BhatApp-1996,BhatApp-2016,BhatApp-2017} or the Chernoff distance~\cite{Chernoff-2013,ChernoffIG-2017,Chernoff-2017} for exponential families (e.g., Gaussian/multinoulli): This highlights the  important role in disguise of the equivalent skew Jensen divergences (see Eq.~\ref{eq:bhat2jensen}).

The paper is organized as follows:
Section~\ref{sec:chorddiv} introduces the novel triparametric family of chord gap divergences that generalizes the skew Jensen divergences (\S\ref{sec:chorddiv-def}), describes several properties (\S\ref{sec:chorddiv-prop}), and deduces a generalization of the statistical Bhattacharyya distance (\S~\ref{sec:gen}).
Section~\ref{sec:clustering} considers the calculation of the centroid (\S\ref{sec:centroid}) for the chord gap divergences, and report probabilistic guarantee of
 the $k$-means++ seeding (\S\ref{sec:kmpp}) by highlighting the Taylor-Lagrange remainder forms of those divergences.

\section{The chord gap divergence}\label{sec:chorddiv}

\subsection{Definition}\label{sec:chorddiv-def}

\begin{figure} 
\centering
\includegraphics[width=0.7\textwidth]{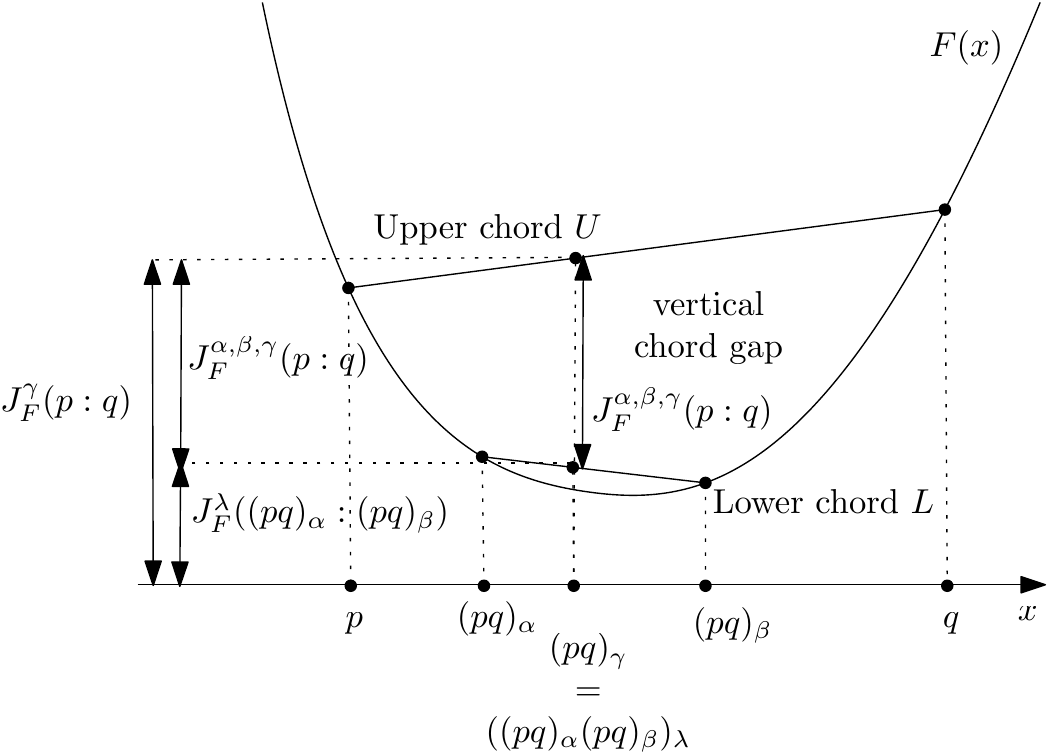}
\caption{The triparametric chord gap divergence: The vertical distance beetween the upper chord $U$ and the lower chord $L$ is non-negative and zero iff. $p=q$.\label{fig:chorddiv}}
\end{figure}

Let $F:\calX\rightarrow \bbR$ be a {\em strictly convex} function.
For $\alpha,\beta\in (0,1)$ with $\alpha\not=\beta$ and $\alpha\beta<1$, 
the chord
\begin{equation}
L=[((pq)_\alpha,F((pq)_\alpha))((pq)_\beta,F((pq)_\beta))],
\end{equation} 
is  below the distinct chord 
\begin{equation}
U=[(p,F(p))(q,F(q))].
\end{equation} 
Thus we can define a divergence as the {\em vertical gap} between these two U/L chords for a given coordinate 
$x\in [(pq)_\alpha,(pq)_\beta]$:
\begin{equation}
\boxed{J_F^{\alpha,\beta,\gamma}(p:q)=(F(p)F(q))_\gamma -(F((pq)_\alpha)F((pq)_\beta))_\lambda}
\end{equation}
 such that $((pq)_\alpha(pq)_\beta)_\lambda=(pq)_\gamma$ with $\gamma\in (\alpha,\beta)$ (see Figure~\ref{fig:chorddiv}).
 A calculation shows that:
\begin{equation}
\lambda=\lambda(\alpha,\beta,\gamma)=\frac{\gamma-\alpha}{\beta-\alpha},
\end{equation} 
or that 
\begin{equation}
\gamma=\lambda(\beta-\alpha)+\alpha,
\end{equation} 
for $\lambda\in [0,1]$ ($\gamma\in [\alpha,\beta]$)
when $\alpha\not=\beta$, so that we get :
\begin{eqnarray}
J_F^{\alpha,\beta,\gamma}(p:q) &=& (F(p)F(q))_\gamma -(F((pq)_\alpha)F((pq)_\beta))_{\frac{\gamma-\alpha}{\beta-\alpha}},\quad \gamma\in [\alpha,\beta]\\
&=& (F(p)F(q))_{\lambda(\beta-\alpha)+\alpha} - (F((pq)_\alpha)F((pq)_\beta))_{\lambda},\quad \lambda\in [0,1].
\end{eqnarray}

\subsection{Properties of the chord gap divergence and subfamilies}\label{sec:chorddiv-prop}

We have:
\begin{eqnarray}
J_F^{\alpha,\alpha,\alpha}(p:q)&=&J_F^{\alpha}(p:q),\\
 J_F^{0,1,\gamma}(p:q)&=&J^\gamma(p:q),\\ 
J_F^{\alpha,\beta,\gamma}(q:p)&=& J_F^{1-\alpha, 1-\beta,1-\gamma}(p:q),
\end{eqnarray}
since $\lambda(1-\alpha,1-\beta,1-\gamma)=\frac{\gamma-\alpha}{\beta-\alpha}=\lambda(\alpha,\beta,\gamma)$ using the fact that $(ab)_{1-\delta}=(ba)_\delta$ for $\delta\in [0,1]$. Thus we also have: 
\begin{equation}
J_F^{1-\alpha,1-\alpha,1-\alpha}(p:q)=J_F^\alpha(q:p).
\end{equation}
          
More importantly, we can express the chord gap divergence as the difference of two skew Jensen divergences (Figure~\ref{fig:chorddiv}):
\begin{equation}\label{eq:Jdiff}
J_F^{\alpha,\beta,\gamma}(p:q)=J_F^\gamma(p:q)-J_F^\lambda((pq)_\alpha:(pq)_\beta),
\end{equation}
with $\lambda=\frac{\gamma-\alpha}{\beta-\alpha}$ or $\gamma=\lambda(\beta-\alpha)+\alpha$ for $\lambda\in [0,1]$ and $\gamma\in [\alpha,\beta]$. 
Thus the chord gap divergence can be interpreted as a {\em truncated} skew Jensen divergence: The truncation of the vertical gap measured by 
$J_F^\gamma(p:q)$ from which we remove the vertical gap measured by $J_F^\lambda((pq)_\alpha:(pq)_\beta)$.

A biparametric subfamily $J_F^{\beta,\gamma}$  of $J_F^{\alpha,\beta,\gamma}$ is obtained by setting $\alpha=0$ so that $(pq)_\alpha=p$, so that the two upper/lower chords $L$ and $U$ coincide at extremity $p$:

\begin{eqnarray}
J_F^{\beta,\gamma}(p:q)&=&(F(p)F(q))_{\gamma}-(F(p)F((pq)_\beta))_{\frac{\gamma}{\beta}},\\
&=& \left(\frac{\gamma}{\beta} -\gamma \right) F(p)+\gamma F(q)-\frac{\gamma}{\beta} F((pq)_\beta),\\
&=&  \gamma \left( \left(\frac{1}{\beta} -1 \right) F(p)+ F(q)-\frac{1}{\beta} F((pq)_\beta)\right).
\end{eqnarray}
 
When $\beta=\frac{1}{2}$, we find that $J_F^{\frac{1}{2},\gamma}(p:q)=2\gamma J_F(p:q)$:
\begin{eqnarray}
J_F^{\gamma}(p:q)&=&2\gamma \left(\frac{F(p)+F(q)}{2} -F\left(\frac{p+q}{2}\right) \right)\\
\end{eqnarray}
is the ordinary ($\gamma$-scaled) Jensen divergence.
When $\beta\rightarrow 0$, we have $\lim_{\beta\rightarrow 0} \frac{1}{\gamma} J_F^{\beta,\gamma}(p:q)=B_F(q:p)$ (with $\gamma\in(0,\beta)$)
since $-\frac{1}{\beta} F((pq)_\beta)\simeq -\frac{1}{\beta}-(q-p)^\top \nabla F(p)$ using a first-order Taylor expansion.

We may also consider $\beta=1-\alpha$, and define the biparametric subfamily: 
\begin{eqnarray}
{J'}_F^{\alpha,\gamma}(p:q) &=& (F(p)F(q))_\gamma -(F((pq)_\alpha)F((pq)_{1-\alpha}))_{\frac{\gamma-\alpha}{1-2\alpha}},\quad \gamma\in [\alpha,1-\alpha],\\
&=& (F(p)F(q))_{\lambda(1-2\alpha)+\alpha} - (F((pq)_\alpha)F((qp)_\alpha))_{\lambda},\quad \lambda\in [0,1].
\end{eqnarray}

Chord gap divergences operating on matrix arguments can be obtained by taking  strictly convex matrix generators~\cite{JB-2013} (e.g., $F(X)=-\log\det |X|$ where $|X|$ denotes the determinant of $X$) for symmetric positive definite matrices $X\in \bbP_{++}$, $\bbP_{++}=\{X\ : \ X\succ0 \}$ denote the space of positive definite matrices,   a convex cone. This may be useful in applications based on covariance matrices~\cite{JB-2013} (or correlation matrices). 

\subsection{Generalized Bhattacharrya distances}\label{sec:gen}
The  interpretation given in Eq.~\ref{eq:Jdiff} yields a triparametric family of Bhattacharryya statistical distances~\cite{Bhat-1946} between members $p(x)=p(x;\theta_p)$ and  $q(x)=p(x;\theta_q)$ of the {\it same} exponential family (with a slight abuse of notation
 that $\Bhat_\alpha(\theta_p:\theta_q)=\Bhat_\alpha(p(x;\theta_p):p(x;\theta_q))$):

\begin{equation}
\Bhat_{\alpha,\beta,\gamma}(\theta_p:\theta_q)= \Bhat_\gamma(\theta_p:\theta_q)-\Bhat_\lambda((\theta_p\theta_q)_\alpha:(\theta_p\theta_q)_\beta).
\end{equation}
It follows that:
\begin{eqnarray}\label{eq:gbhatef}
\lefteqn{\Bhat_{\alpha,\beta,\gamma}(\theta_p:\theta_q) =}\\ \nonumber 
&& -\log 
\frac{\int p(x;\theta_p)^{1-\gamma} p(x;\theta_q)^{\gamma} \dx}{
\int p(x;(\theta_p\theta_q)_\alpha)^{1-\lambda} p(x;(\theta_p\theta_q)_\beta)^{\lambda} \dx}.
\end{eqnarray}
Note that when $\alpha=\beta$, we have:
\begin{equation}
p(x;(\theta_p\theta_q)_\alpha)^{1-\lambda} p(x;(\theta_p\theta_q)_\beta)^{\lambda}=p(x;(\theta_p\theta_q)_\alpha),
\end{equation} 
and therefore the denominator becomes $\int p(x;(\theta_p\theta_q)_\alpha)\dx=1$, and we recover the skew Bhattacharryya distance, as expected.

We shall extend the generalized Bhattacharrya divergence of Eq.~\ref{eq:gbhatef} to arbitrary distributions by
generalizing the notion of {\em interpolated distribution}:
\begin{equation}
p(x;(\theta_p\theta_q)_\delta)=\Gamma_\delta(p(x;\theta_p),p(x;\theta_q)).
\end{equation}
When $\delta$ ranges from $0$ to $1$, we obtain a {\em Bhattacharyya arc} linking distribution $p(x;\theta_p)$ 
to distribution $p(x;\theta_q)$ (the arc is called an exponential or $e$-geodesic in information geometry~\cite{IG-2016}).
We define:
\begin{equation}
\Gamma_\delta(p(x),q(x)) =\frac{p(x)^{1-\delta}q(x)^{\delta}}{Z_\delta(p(x):q(x))},
\end{equation}
with
\begin{equation}
Z_\delta(p(x):q(x))=\int p(x)^{1-\delta}q(x)^{\delta}\dnu(x).
 \end{equation}
Note that we need the integral to converge properly in order to define $\Gamma_\delta(p(x),q(x))$. 
This always holds  for distributions belonging to the same exponential families
since  $(\theta_p\theta_q)_\delta$ is guaranteed to belong to the natural parameter space, and
\begin{equation}
Z_\delta(p(x;\theta_p):p(x;\theta_q))=\exp(-J_F^\delta(\theta_p:\theta_q)).
\end{equation}

By extension, the triparametric Bhattacharryya distance can be defined by:
\begin{eqnarray}
\lefteqn{\Bhat^{\alpha,\beta,\gamma}(p(x):q(x)) =}\nonumber\\ 
&& -\log \left( \frac{\int p(x)^{1-\gamma} q(x)^{\gamma} \dnu(x)}{
\Gamma_\alpha(p(x),q(x))^{1-\gamma} \Gamma_\beta(p(x),q(x))^{\gamma}} \right).
\end{eqnarray}
Thus we explicitly define the generalized Bhattacharrya distance by:
{ 
\begin{eqnarray*}
\lefteqn{\Bhat^{\alpha,\beta,\gamma}(p(x):q(x))=}\nonumber\\
&& -\log \left( 
\frac{\int p(x)^{1-\gamma} q(x)^{\gamma} \dnu(x)}{
\int \left(\frac{p(x)^{1-\alpha}q(x)^{\alpha}\dnu(x)}{\int p(x)^{1-\alpha}q(x)^{\alpha}\dnu(x)}\right)^{1-\lambda} 
\left(\frac{p(x)^{1-\beta}q(x)^{\beta}\dnu(x)}{\int p(x)^{1-\beta}q(x)^{\beta}\dnu (x)} \right)^{\lambda} \dnu(x)
}\right).
\end{eqnarray*}
}

Notice that when $\alpha=\beta$, for any $\lambda\in [0,1]$, the denominator collapses to one, and we find
 that $\Bhat_{\alpha,\beta,\gamma}(p(x):q(x)) = \Bhat_{\alpha}(p(x):q(x))$, as expected.

For multivariate gaussians/normals belonging to the family $\{\calN(\mu,\Sigma)\ :\ \mu\in\bbR^d,\Sigma\in\bbP^d_{++}\}$, we have 
the natural parameter~\cite{SMEF-2011} 
\begin{equation}
\theta=(v,M)=(\Sigma^{-1}\mu,-\frac{1}{2}\Sigma^{-1}),
\end{equation} 
and the 
cumulant function:
\begin{equation}
F(v,M)=\frac{d}{2}\log 2\pi-\frac{1}{2}\log |-2M|-\frac{1}{4}v^\top M^{-1}v,
\end{equation} 
that can also be expressed in the usual parameters:
\begin{equation}
F(\mu,\Sigma)=\frac{1}{2}\log (2\pi)^d|\Sigma|+\frac{1}{2}\mu^\top\Sigma^{-1}\mu.
\end{equation}
We have:
\begin{equation}
(\theta_p\theta_q)_\delta=((1-\delta)\Sigma_p^{-1}\mu_p+\delta\Sigma_q^{-1}\mu_q,-\frac{1-\delta}{2}\Sigma_p^{-1}-\frac{\delta}{2}\Sigma_q^{-1}),
\end{equation} 
so that we get~\cite{hero-2001}:
\begin{equation}
J_F^\alpha(p(x;\mu_p,\Sigma_p):p(x;\mu_q,\Sigma_q))=\frac{\alpha(1-\alpha)}{2}\Delta\mu^\top((1-\alpha)\Sigma_p+\alpha\Sigma_q)^{-1}\Delta\mu
+
\frac{1}{2}\log\frac{|(1-\alpha)\Sigma_p+\alpha\Sigma_q|}{|\Sigma_p|^{1-\alpha}|\Sigma_q^\alpha|},
\end{equation}
where $|.|$ denotes the matrix determinant and $\Delta\mu=\mu_q-\mu_p$.
This gives a closed-form formula for $\Bhat^{\alpha,\beta,\gamma}$ for multivariate Gaussians.
See~\cite{clusteringmvn-2009} for applications clustering multivariate normals.


\section{Centroid-based clustering}\label{sec:clustering}
Bhattacharrya clustering is often used in statistical signal processing, information fusion, and machine learning (see~\cite{BhatApp-1996,BhatApp-1979,BhatApp-2006,BhatKernel-2009,BhatC-2014,BhatC-2015} for some illustrative examples).
Popular clustering algorithms are {\em center-based clustering}, where each cluster stores a prototype (a representative element of the cluster), and each datum is assigned to the cluster with the closest prototype with respect to a distance function. The cluster prototypes are then updated, and the algorithm iterates until (local) convergence.
This scheme includes the $k$-means and the $k$-medians~\cite{kmeanscoresets-2004}. 
Lloyd $k$-means heuristic updates the prototype $c$ of a cluster $X$ by choosing its center of mass $c=\frac{1}{|X|}\sum_{x\in X} x$ that minimizes the cluster variance: $\min_c \sum_{x\in X} \|x-c\|^2$ (this holds for any Bregman divergence too~\cite{Bregman-2005}).

\subsection{Chord gap divergence centroid}\label{sec:centroid}

We extend $k$-means for a weighted point set: 
\begin{equation}
\calP=\{(w_1,p_2), \ldots, (w_n,p_n)\},
\end{equation}
with $w_i>0$ and $\sum_i w_i=1$, using the chord gap divergence by solving the following minimization problem: 
\begin{equation}
\min_x E(x)=\sum_{i=1}^n  w_i J_F^{\alpha,\beta,\gamma}(p_i:x).
\end{equation}

By expanding the chord gap divergence formula and removing all terms independent of $x$, we obtain an equivalent minimization problem as a difference  of convex function programming~\cite{DC-2005}: 
\begin{equation}
\min_x E(x)=\min_x A(x)-B(x),
\end{equation} 
with 
\begin{eqnarray}
A(x)&=&\sum_{i=1}^n (F(p_i)F(x))_\gamma, \\ 
B(x)&=&\sum_{i=1}^n (F((p_ix)_\alpha)F((p_ix)_\beta))_\lambda,
\end{eqnarray}
 both strictly convex functions.
It follows a concave-convex procedure~\cite{CCCP-2002} (CCCP) solving locally $\min_x A(x)-B(x)$: 
initialize $x_0=p_1$ and then iteratively update as follows:
\begin{equation}
\nabla A(x_{t+1})=\nabla B(x_{t}).
\end{equation}
When the reciprocal gradient $\nabla A^{-1}$ (such that $\nabla A^{-1}(\nabla A(x))=x$) is available in closed form, we end up with the following update:
\begin{equation}
x_{t+1}=\nabla A^{-1}(\nabla B(x_{t})).
\end{equation}
Since we have 
\begin{eqnarray}
\nabla A(x)&=&n\gamma \nabla F(x),\\  
\nabla B(x)&=&\sum_i (1-\lambda)\alpha\nabla F(F((p_ix)_\alpha) +\lambda\beta \nabla F((p_ix)_\beta),
\end{eqnarray} 
the update rule is
{%
\begin{eqnarray*}
\lefteqn{x_{t+1}=}\nonumber\\
&&\nabla F^{-1}\left(
\frac{1}{\gamma} \sum_i w_i\left( (1-\lambda)\alpha\nabla F((p_ix_t)_\alpha) +\lambda\beta \nabla F\left((p_ix_t)_\beta\right)\right) 
\right).
\end{eqnarray*}
}

When $\alpha=\beta=\gamma$, we find the simplified update rule:
\begin{equation}
x_{t+1}=\nabla F^{-1}\left(\sum_i w_i \nabla F((p_ix_t)_\alpha\right),
\end{equation}
corresponding to the skew Jensen divergences~\cite{Jensen-2011}. 
Note that it is enough to improve iteratively the prototypes to get a variational Lloyd's $k$-means~\cite{tJ-2015} that guarantees monotone convergence to a (local) optimum.

\subsection{Performance analysis of $k$-means++}\label{sec:kmpp}

For high-performance clustering, one may use $k$-means++~\cite{kmpp-2007} that is a guaranteed probabilistic initialization of the cluster prototypes.
For an arbitrary dissimilarity function $D(\cdot:\cdot)$, the cost function of $k$-means (also called potential function or energy function) for a weighted point set $\calP$ with cluster center set $\calC$ is defined by
$$
\Phi(\calC)=\sum_{i=1}^n w_i \min_{c\in\calC} D(p_i:c).
$$ 
Let $\Phi^*=\min_{\calC\ :\ |C|=k} \Phi(\calC)$ denote the optimal cost function.
A heuristic $H$  delivering a cluster center set $\calC_H$ is said $\kappa$-competitive when $\frac{\Phi(\calC_H)}{\Phi^*}=\kappa$.
$\kappa$ is termed the competitive ratio~\cite{kmpp-2007}.
To get an expected competitive ratio of $2U^2(1+V)(2+\log k)$~\cite{tJ-2015}, we need to upper bound:
\begin{itemize}
\item $U$ such that the  divergence $D=J_F^{\alpha,\beta\gamma}$ satisfies the $U$-triangular inequality $D(x:z)\leq U(D(x:y)+D(y:z))$, and
\item $V$ such that the divergence satisfies the symmetric inequality $D(y:x)\leq V D(x:y)$.
\end{itemize}

The proof  follows the proof reported in~\cite{tJ-2015} for total Jensen divergences once we can express the divergences in their 
Taylor-Lagrange remainder forms:
\begin{equation}
D(p:q)=(p-q)^\top H_D(p:q) (p-q),
\end{equation} 
where $H_D(p:q)\succ 0$.
For example, the Taylor-Lagrange remainder form of the Bregman divergence~\cite{Mixed-2008} is obtained from a first-order Taylor expansion with the exact Lagrange remainder:
\begin{equation}\label{eq:tlbreg}
B_F(p:q)=\frac{1}{2} (p-q)^\top \nabla^2F(\xi) (p-q),
\end{equation} 
for some $\xi\in [p,q]$. This expression can be interpreted as a squared Mahalanobis distance:
\begin{equation}
M_Q(p,q)=(p-q)^\top Q (p-q),
\end{equation} 
with precision matrix $Q=\frac{1}{2}\nabla^2F(\xi)\succ 0$ depending on $p$ and $q$. 
Any squared Mahalanobis distance satisfies $U=2$ (see~\cite{Ackermann-2010}) and $V=1$, and can be interpreted as a squared norm-induced distance:
\begin{equation}
M_Q(p,q)=\|Q^{\frac{1}{2}}(p-q)\|_2^2.
\end{equation}

We report the Taylor-Lagrange remainder form of the skew Jensen divergences:
There exists $\xi_1,\xi_2\in [p,q]$, such that the skew Jensen divergence can be expressed as
$J_F^\alpha(p:q)=(p-q)^\top H_F^\alpha(p:q) (p-q)$,
with 
\begin{equation}
H_F^\alpha(p:q)=\frac{1}{2}\alpha(1-\alpha)(\alpha\nabla^2 F(\xi_1)+(1-\alpha)\nabla^2 F(\xi_2)).
\end{equation}
 
The proof relies on introducing the skew Jensen-Bregman (JB) divergence~\cite{Jensen-2011} defined by
\begin{equation}
\JB_F^\alpha(p:q) =(1-\alpha)B_F(p:(pq)_\alpha)+\alpha B_F(q:(pq)_\alpha),\\
\end{equation}
and observing the $\JB_F^\alpha(p:q) =J_F^\alpha(p:q)$
since $p-(pq)_\alpha=\alpha(p-q)$ and $q-(pq)_\alpha=(1-\alpha)(q-p)$ (and therefore the   $\nabla F((pq)_\alpha)$-terms cancel out). 
Then we apply the Taylor-Lagrange remainder form of Bregman divergences of Eq.~\ref{eq:tlbreg} to get the result.
Notice that when $\alpha\rightarrow 0$ or $\alpha\rightarrow 1$, the scaled skew Jensen difference tend to Bregman divergences, and we have
$\lim_{\alpha\rightarrow 1} \frac{J_F^{(\alpha)}(p:q) }{\alpha(1-\alpha)} =  \frac{1}{2}(p-q)^\top \nabla^2 F(\xi_1) (p-q)=B_F(p:q)$
for $\xi_1\in [p,q]$, and $\lim_{\alpha\rightarrow 0} \frac{J_F^{(\alpha)}(p:q)}{\alpha(1-\alpha)}=   \frac{1}{2}(p-q)^\top \nabla^2 F(\xi_2) (p-q)= B_F(q:p)$
for $\xi_2\in [p,q]$
, as expected.

Using expression of Eq.~\ref{eq:Jdiff} for the chord gap divergence, and 
 the fact that $(pq)_\alpha-(pq)_\beta=(\alpha-\beta)(q-p)$, we get the Taylor-Lagrange form
 of the chord gap divergence $J_F^{\alpha,\beta,\gamma}=(p-q)^\top H_F^{\alpha,\beta,\gamma}(p:q) (p-q)$
with
{
\begin{eqnarray}
H_F^{\alpha,\beta,\gamma}(p:q)&=& \frac{1}{2}\gamma(1-\gamma)\nabla^2 F(\xi')-\frac{1}{2}\lambda(1-\lambda)(\alpha-\beta)^2\nabla^2 F(\xi''),\\
&=& \frac{1}{2}\left(\gamma(1-\gamma)\nabla^2 F(\xi')-(\gamma-\alpha)(\gamma-\beta)\nabla^2 F(\xi'')\right),
\end{eqnarray}
}
for $\xi',\xi''\in\calX$.

An alternative proof considers the Taylor first-order expansion of  $F$ with exact Lagrange remainder:
\begin{equation}
F(x)=F(a)+(x-a)^\top \nabla F(a)+\frac{1}{2} (x-a)^\top \nabla^2 F(\xi) (x-a),\quad \xi\in [a,x].
\end{equation}
Therefore  we get the following Taylor   expansions with exact Lagrange remainders:

\begin{eqnarray}
F(p) &=& F((pq)_\alpha) + \alpha(p-q)^\top \nabla F((pq)_\alpha)+  \frac{1}{2}\alpha^2(p-q)^\top \nabla^2 F(\xi_1) (p-q), \quad \xi_1\in (p,(pq)_\alpha),\\
F(q) &=& F((pq)_\alpha) + (1-\alpha)(q-p)^\top \nabla F((pq)_\alpha)+  \frac{1}{2}(1-\alpha)^2(q-p)^\top \nabla^2 F(\xi_2) (q-p),
  \xi_2\in (q, (pq)_\alpha).
\end{eqnarray}

Multiplying the first equation by $1-\alpha$ and the second equation by $\alpha$ and summing up, we obtain:

\begin{equation}
(1-\alpha)F(p)+\alpha F(q)= F((pq)_\alpha)+ \frac{1}{2}\alpha^2(1-\alpha)(p-q)^\top  \nabla^2 F(\xi_1)(p-q)+  \frac{1}{2}(1-\alpha)^2\alpha \nabla^2 F(\xi_2)(p-q),
\end{equation}
since the gradient terms cancel out, and we get:
\begin{eqnarray}
J_F^{(\alpha)}(p:q)&=&(1-\alpha)F(p)+\alpha F(q)-F((pq)_\alpha),\\
&=& \frac{1}{2}\alpha(1-\alpha)(p-q)^\top \left(\alpha \nabla^2 F(\xi_1)+ (1-\alpha) \nabla^2 F(\xi_2)\right) (p-q).
\end{eqnarray}

Thus it follows the Taylor-Lagrange remainder form of skew Jensen divergences: 

\begin{equation}
\boxed{J_F^{(\alpha)}(p:q) =  \frac{\alpha(1-\alpha)}{2}(p-q)^\top (\alpha \nabla^2 F(\xi_1)+(1-\alpha) \nabla^2 F(\xi_2)) (p-q).}
\end{equation}

When dealing with a finite (weighted) point set $\calP$, let 
\begin{equation}
\rho=\frac{\sup_{\xi',\xi'',p,q\in\co(\calP)} \|(\nabla^2 F(\xi'))^{\frac{1}{2}}(p-q)\|}{\inf_{\xi',\xi'',p,q\in\co(\calP)} \|(\nabla^2 F(\xi''))^{\frac{1}{2}}(p-q)\|}<\infty,
\end{equation} 
where $\co(\calP)$ denotes the convex closure of $\calP$.
Then it comes that $U=O_\rho(1)$ and $V=O_\rho(1)$ so that $k$-means++ probabilistic seeding is $\bar{O}_\rho(\log k)$ competitive for the chord gap divergence.

\section{Concluding remarks}\label{sec:concl}
We introduced the chord gap divergence as a generalization of the skew Jensen divergences~\cite{Jensen-2011,JB-2013}, studied its properties
and obtained a generalization of the skew Bhattacharrya divergences.
We showed that the chord gap divergence  centroid can be obtained using a convex-concave iterative procedure~\cite{Jensen-2011}, and analyzed the $k$-means++~\cite{kmpp-2007} performance by giving the Taylor-Lagrange forms of the skew Jensen and chord gap divergences.
We expect our contributions to be useful for the signal processing, information fusion and machine learning communities where the Bhattacharrya~\cite{BhatG-2008,BhatG-Clustering-2016} or Chernoff information~\cite{Bhat-1982,Chernoff-2013} is often used. 
In practice, the triparametric chord gap divergence shall be tuned according to the application at hand (and the dataset for supervised tasks using cross-validation for example).
  
Public Java\texttrademark{} source code is available for reproducible research:\\
 \centerline{\url{http://www.lix.polytechnique.fr/~nielsen/CGD/}}


\end{document}